# A hybrid IndRNNLSTM approach for real-time anomaly detection in software-defined networks

***Sajjad Salem[a], Salman Asoudeh[b]***

[a]salemsajjad@aut.ac.ir
[b]s.asouded@velayat.ac.ir

ABSTRACT

Anomaly detection in SDN using data flow prediction is a difficult task. This problem is included in the category of time series and regression problems. Machine learning approaches are challenging in this field due to the manual selection of features. On the other hand, deep learning approaches have important features due to the automatic selection of features. Meanwhile, RNN-based approaches have been used the most. The LSTM and GRU approaches learn dependent entities well; on the other hand, the IndRNN approach learns non-dependent entities in time series. The proposed approach tried to use a combination of IndRNN and LSTM approaches to learn dependent and non-dependent features. Feature selection approaches also provide a suitable view of features for the models; for this purpose, four feature selection models, Filter, Wrapper, Embedded, and Autoencoder, were used. The proposed IndRNNLSTM algorithm, in combination with Embedded, was able to achieve MAE=1.22 and RMSE=9.92 on NSL-KDD data.

## 1. Introduction:

SDN anomaly detection mainly detects and analyzes abnormal traffic by extract- ing statistical characteristics of the traffic. A complete understanding of the concept of symmetry is essential in detecting anomalies and reducing anomalies. However, basic
information about network traffic is easily lost, and setting up dynamic network con- figuration becomes increasingly complex. Therefore, we need a proposed system that
uses the advantages of SDN architecture and detects anomalies in real-time using ma- chine learning algorithms within a few milliseconds and alerts a user about an attack to take necessary actions[1]. An intrusion detection system (IDS) is a network security device that monitors network and/or system activities for malicious or unwanted behavior[3]. There are different types of intrusion detection systems:

- network-based intrusion detection: Network-based IDS (NIDS) looks for at- tack signatures in network traffic [4]. Typically, a network adapter running in non-stop mode monitors and analyzes all traffic as it travels across the network. The attack detection module uses network packets as a data source.
- Host-based intrusion detection: host-based IDS (HIDS) look for attack signatures in host log files [5].
- Statistical abnormality diagnosis: Statistical anomaly detection (or behavior-based detection) uses statistical techniques to identify potential intrusions. First," normal" behavior is defined as a baseline. During operation, statistical analysis of the monitored data is performed and the deviation from the baseline is measured. If the threshold is exceeded, an alarm will be issued. This type of IDS does not need to know the security vulnerabilities in a particular system. The base-line defines normality. Therefore, there is an opportunity to detect new attacks without the need to update the knowledge base. On the other hand, anomaly detection only detects anomalies. Suspicious behavior is not always defined as an intrusion. For example, a number of failed login attempts may be due to an attack or password being forgotten by the administrator.

Time series anomaly detection depends on reliable, consistent and complete data. Poor data quality can lead to inaccurate forecasting results or prevent anomalies from being detected altogether. Factors that can affect data quality include [2]:

- Missing values
- Outliers
- Seasonality
- Trends

**2. Background Research**

Intrusion is" any set of actions that attempt to compromise the integrity, confidentiality or availability of resources" [6]. Intruders can gain access to a system by exploiting weaknesses due to design and programming errors or simply through the negligence of system administrators. IDS is a set of software and hardware that tries to detect computer attacks by examining various data audits and sorting them as normal or malicious activities [7]. The robustness of an IDS primarily depends on the strength of the intrusion detection technology used, and over the years, researchers have tried to find the best combination of accuracy and low modelling time. The following discusses some of the most important works done for NSL-KDD. A model based on Stacking Random Forest and KNN approaches was proposed in [8] for the NSL-KDD dataset. This approach combined two basic machine learning models and achieved 81.67 accuracy on the test data in the best case. The combination of LSTM and a fully connected model was used in [9]. This approach was tried first to extract the primary features using the LSTM network and then to use the fully connected network to learn the extracted features. This approach reached a maximum accuracy of 99.95 Combining LSTM and Autoencoder in [10] obtained competitive results in NSL-KDD. Their method has a 6-layer Autoencoder (AE) model with LSTM, which is effective in anomaly detection. To avoid bias in the model that occurs from unbalanced data in the NSL-KDD dataset, they used a standard scaler in their AE-LSTM model to remove outliers from the input. AE-LSTM uses the best reconstruction performance, which is very important in detecting the normality or abnormality of network traffic. Their proposed model has the highest accuracy compared to other methods with f1-score micro and weight at 98.69% and 98.70% for five classes in detection methods, and two classes (Malicious, Normal) with f1-score micro and weight at 98.78% and 98.78% got.

Also, researchers in [11] presented an IDS based on deep learning using RNN. In this research, the authors used Simple RNN. In their proposed framework, a training set was entered into a data processing block, which converted classified data into numerical inputs. In addition, all inputs were normalized using a scaling function. In addition, the data processing block sent information to the training block to train and build the model. The dataset used in this study was the NSL-KDD dataset. To evaluate the performance, the authors considered the accuracy obtained through the test data as the primary criterion for an optimal model. The results showed that RNN-IDS achieved a test accuracy of 83.28% for the binary classification scheme. In contrast, RNN-IDS achieved an accuracy of 81.29% for the 5-way classification task. This research did not implement any feature reduction techniques that could increase the performance of RNN-IDS or reduce training and testing time.

In [12], the authors presented an IDS based on LSTM combined with multivariate correlation analysis (MCA). MCA-LSTM used Information Gain (IG) as a feature se- lection approach. First, the model selects a subset of features. In addition, the selected subsets were converted into a triangle area map matrix (TAM). Finally, TAM was used by the LSTM algorithm to predict penetration. In order to measure the performance of their model, the authors used the NSL-KDD and UNSW-NB15 datasets. Experimental results showed that MCA-LSTM achieved a test accuracy of 82.15% for 5-class classification using NSL-KDD. Although these results were superior to other methods, the authors did not study the effect of dataset size or consider a wide range of performance measures such as F1 score.

In [13], two deep learning techniques based on LSTM and CNN were proposed for intrusion detection. First, the authors introduced an IDS based on the LSTM method. In the second step, researchers presented a Convolutional Neural Network (CNN) together with the LSTM algorithm. The NSL-KDD dataset was used as the basis for their experiments. In the data processing step, the authors used the input scaling method to accommodate changes in feature values. The authors used the accuracy of the test data (KDD Test +) as the main criterion to evaluate the performance of their models. The proposed LSTM achieved 74.770% and 68.780% accuracy for binary and multi-class classification tasks, respectively. CNN-LSTM achieved a test accuracy of 79.37% for the binary classification scheme and 70.13% for the 5-class classification. The authors of this research concluded that implementing a suitable feature extraction method can improve the results obtained from their proposed methods.

**3. Proposed Method**

Due to the automatic selection of features, deep learning approaches have been able to achieve higher accuracy in many time series cases, etc., than traditional ma- chine learning approaches. Selecting manual features is time-consuming, expensive and requires special knowledge, which creates challenges for models. Today, forecasting accuracy is one of the most important factors in choosing forecasting methods. In general, choosing the most effective forecasting method is a very difficult task, and this has caused many researchers to integrate linear and non-linear methods in order to achieve more accurate and complete results. Approaches based on recurrent neural networks and deep learning have also been proposed for this purpose. In fact, the use of deep learning approaches has largely solved these challenges. Among the results derived from the methods based on neural networks, RNNs have been able to obtain better results in the tasks of time series regression and time series classification by extracting time series features than other approaches. LSTM is an improved version of recurrent networks, but recently, the GRU algorithm has been proposed to improve these net- works. These networks and approaches also have problems. One of the most important problems is that in recurrent neural networks such as LSTM and GRU, temporal dependencies are considered long and non-independent. This problem affects the overall accuracy of the model because the use of independent dependencies can increase the accuracy of the model and reduce complex dependencies. One of the solutions that can be used instead of LSTM and GRU is the IndRNN approach [14]. This approach tries to learn inputs in a non-dependent manner. This approach has advantages over LSTM and GRU networks, the most important of which are [14]:

1. It can control the problem of vanishing and exploding gradients well.
2. Short-term memory blocks can learn longer sequences with IndRNN.
3. IndRNN can be trained better with unstructured functions such as ReLu.
4. The existence of several layers of IndRNN in the form of a stack can be trained better.

The block diagram of the suggested approach is in figure 1.

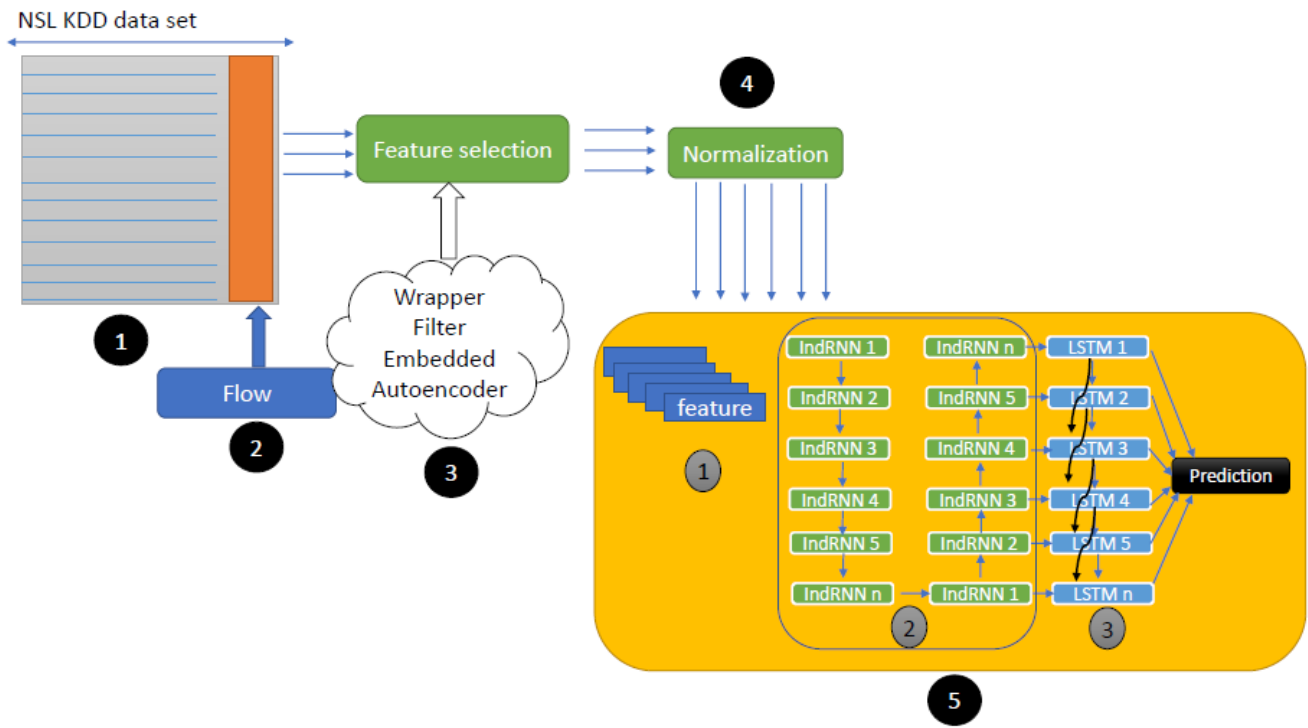

**Figure 1: suggested method for anomaly prediction.**

**Feature selection:** As with the first proposed approach, the described feature selections are tested. Three general feature selection approaches are used to derive a model according to the relationship between feature selection algorithms and the inductive learning method. These items are described below:

- Filters
- Wrappers
- Embedded

Random forests are one of the most popular machine-learning algorithms. They are successful because they generally provide good predictive performance, poor fit, and easy interpretability. This interpretability is provided because it is simple to obtain the importance of each variable in the decision tree. In other words, it is easy to calculate the contribution of each variable in decision-making. Feature selection using random forest is included in the category of embedded methods. Embedded quality methods combine filtering and packing methods. They are implemented by algorithms that have their own internal feature selection methods. Some of the advantages of embedded methods are [15]:

- They are very accurate.
- They generalize better.
- They can be interpreted.

The feature selection process is embedded in the learning or model-building phase in the embedded method. It is computationally cheaper than the packing method, and there is less chance of overfitting [15]. Unlike the filter and packer classes, this technique does not separate the learning from the selection part. It uses an internal classification algorithm for the search process. The goal is to reduce the computational time spent classifying different subsets in the packing technique. It's similar to a wrapper but usually faster because it doesn't check all possible combinations like a wrapper. However, according to the classification model, this approach can interact less with the classification model. The table below shows the details of these models. Each of these methods can lead to different results in predictions, so it is tried to use all these models.

**Normalization:** Maximin normalization is used in this step. The relationship of this normalizer is as follows:

$$X_{new} = \frac{X - \mu}{Max - min}$$

which subtracts the average from each data and divides it by the range of its changes.

**Forming sliding windows:** In this step, sliding windows are formed for forecasting. The figure 2 shows how this process works. For example, for forecasting at time X11, all previous values from time X1 to X10 are considered as input, and similarly, for forecasting X12, previous currency values from time X2 to X11 are considered as input. This action is used for all data, including training and test data.

**Separation of the research data set into training and test sets:** In this step, the research data set is separated into two sets of training and test. The training data set is used to learn the model, and the test data set is used to evaluate the model. The amount of training data will be 20% to 80%.

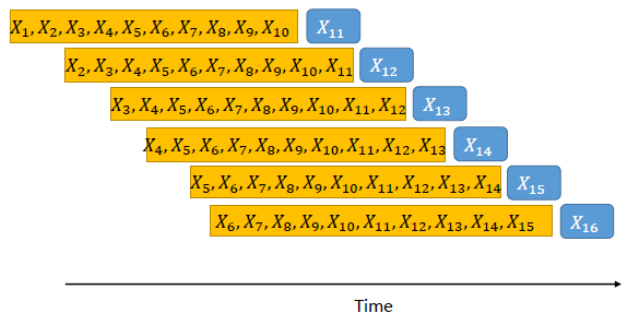

Figure 2: Sliding window

Sending inputs to the IndRNN layer: Independent Recurrent Neural Network (IndRNN) was proposed in [14] as a main component of RNN. It is as follows:

$$h_t = \sigma(W_{x_t} + u \odot h_{t-1} + b)$$

5$x_t \in R^M$ and $h_t \in R^N$ are the input stated and hidden stated in time step $t$, respectively. $w \in R^{N*M}, u \in R^N$, and $b \in R^N$ respectively show the current input, return input, and bias weights; 0 represents the Hadamard product, σ is an essential activation function like ReLu, sigmoid, tanh, or etc. which is ReLu in this paper. Each neuron of IndRNN in a layer is independent of others, and the correlation between neurons is checked by superimposing two or more layers. Since the neurons are independent, the gradient propagation over time can be calculated for each neuron separately. Since the neurons are independent, the gradient propagation over time can be calculated for each neuron separately. For the n − th neuron, the gradient is calculated as follows:

$$\frac{\delta j_{n,T}}{\delta h_{n,t}} = \frac{\delta j_{n,T}}{\delta h_{n,t}} * u_n^{T-1} \prod_{k=T}^{T-1} \sigma'_{n+k+1}$$

Please refer to [14] for more details on the above derivation. This network can be used as a stack. In the proposed approach, it is tried to use both combinations of stack and bidirectional. This part of the network is responsible for feature extraction. The output of this layer is sent as hidden vectors to an LSTM layer. The next step in the proposed method is to send the output of the IndRNN layer as an input to the LSTM layer. A standard LSTM cell contains three gates: the forgetting gate(ft), which determines how much of the previous data will be forgotten; An input gate(it) that evaluates the information to be written to the cell memory. Furthermore, the output gate(ot) that determines how to calculate the output from the current information [3]:

$$f_t = \sigma(Wei_f\, x_t + Uei_f\, h_{t-1} + bei_f)$$
$$it = \sigma(Wei_i\, x_t + Uei_i h_t - 1 + bei_i)$$
$$o_t = \sigma(Wei_o\, x_t + Uei_o h_t - 1 + bei_o)$$
$$c_t = ft \circ ct - 1 + it \circ (Wei_c\, x_t + Uei_c h_t - 1 + bei_c)$$
$$h_t = o_t \circ \tau(ct)$$

where n size of input, m size of cell state and output, $x_t$ input vector(time t, size $n * 1$), ft forget gate vector($m * 1$), it input gate vector($m * 1$), $O_t$ output gate vector(size=$m * 1$), $h_t$ output vector (size=$m * 1$), $c_t$ cell state vector(size= $m * 1$ ), $[Wei_f, Wei_i, Wei_o, Wei_c]$ input gate weight matrices $(size = m * n)$, $[Uei_f, Uei_i, Uei_o, Uei_c]$ output gate weight matrices (size= $m * m$ ), $[bei_f, bei_i, bei_o, bei_c]$ bias vectors(size=m * 1), σ logistic sigmoid activation function, and τ tanh activation function.

Sending the output to the fully connected layer: In the last layer of the proposed method, there are fully connected layers that convert two-dimensional feature maps into one-dimensional feature vectors to continue the feature display process. On the other hand, it can be said that this layer is responsible for calculating the flow. Fully connected layers behave like their counterparts in traditional neural networks and contain approximately 90% of the parameters of a deep belief network. The fully connected layer allows us to represent the output of the network in the form of a vector of a specified size. This vector can be used to classify objects or continue the processing process. One of the effective parameters in this layer is the activator function. Acti- vation functions are essential components of artificial neural networks that are used to provide non-linear mapping of input data and are usually applied to all neurons in a single hidden layer.

## 4. results

MAE and RMSE were used to evaluate the results. Table 4 shows the results of different approaches. We analyze the analysis of these approaches in three different categories. In the first step, we will examine the approach in [16] studies. In the second step, we discuss approaches based on machine learning. In the last stage, we examine the approaches based on deep learning.

Table 1: obtained result by different approaches.

| Model | MAE | RMSE |
|---|---|---|
| Linear Regression[16] | - | 24.221 |
| ARIMA (without windowing)[16] | - | 58.208 |
| ARIMA (with windowing)[16] | - | 20.717 |
| Prophet framework[16] | - | 23.497 |
| SVR Regression | 33.02 | 181.71 |
| KNN Regression | 19.08 | 138.14 |
| Decision Tree | 14.47 | 120.30 |
| Random Forest | 13.05 | 114.23 |
| AdaBoost(ADA) | 32.76 | 181.01 |
| Gradient Boosting(GRAD) | 21.75 | 147.47 |
| Gaussian NB(NB) Regression | 156.72 | 395.88 |
| Linear Discriminant Analysis(LDA) Regression | 32.92 | 181.45 |
| Quadratic Discriminant Analysis(QDA) | 31.158 | 55.81 |
| MLP | 15.74 | 125.49 |
| LSTM | 3.30 | 12.13 |
| GRU | 3.34 | 12.99 |
| IndRNN | 1.74 | 13.19 |
| IndRNN LSTM | 1.4 | 11.96 |

In [16], the authors examined four different approaches to forecasting. Linear regression with a windowing approach has RMSE=24. MAE values for this approach are not provided. On the other hand, combined with the sliding window approach, ARIMA achieved a 26% reduction in error values compared to the simple ARIMA model. Also, among the approaches evaluated in [16], the Prophet framework achieved RMSE=23.497 In the second step, machine learning approaches were evaluated for this problem. These algorithms are based on manual feature selection that tries to map in- puts to outputs to a maximum of one intermediate layer. Feature selection is a challenge in these approaches. For this purpose, nine different approaches were investigated. Initially, SVR was able to achieve MAE=33.02 and RMSE=181.71 on these data. KNN model performed better than SVR and reached MAE=19.08 and RMSE=138.14. The Decision Tree approach reached more acceptable results than SVR and KNN. This approach achieved MAE=14.47 and RMSE=120.30.

Random Forest model was able to reach MAE=13.05 and RMSE=114.23 on these data. This approach worked better than the Decision Tree. ADA was another approach that was evaluated on these data. This approach was able to reach MAE=32.76 and RMSE=181.01. GRAD and NB approaches reached MAE of 21.71 and 156.72, respectively. The two approaches, LDA and QDA obtained close results regarding MAE. These two approaches achieved MAE=32.92 and MAE=31.158, respectively. In terms of RMSE, the QDA approach has performed better. The MLP approach achieved RMSE=125.49 in these data. Traditional machine learning approaches have generally obtained weaker results than[16] approaches. The best RMSE obtained in machine
approaches was 55.81, corresponding to the QDA model. Meanwhile, the best RMSE obtained by [16] approaches equals 20.717, which is related to the ARIMA model (with windowing). Deep learning approaches are applied to the data in the third stage of the proposed models. These approaches try to map inputs to outputs with more than several layers. Different regression models were applied to the data.

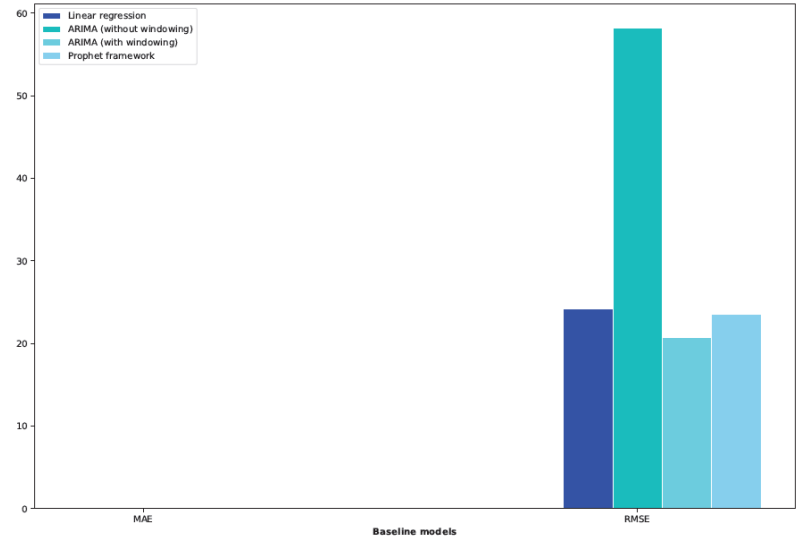

Figure 3: Baseline Model

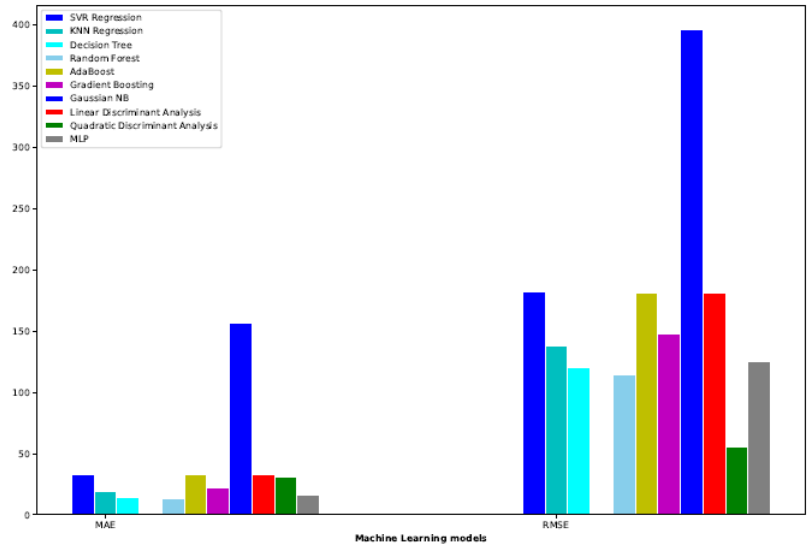

Figure 4: Machine learning Model

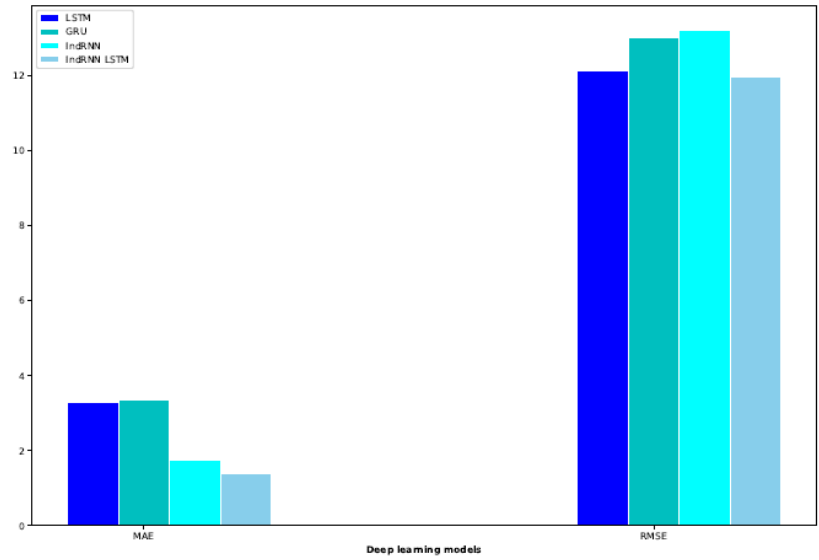

**Figure 5: Deep learning**

The LSTM approach reached MAE=3.30 and RMSE=12.13 on the target data. GRU approach obtained weaker results than LSTM; this approach could reach MAE=3.34 and RMSE=12.99. IndRNN approach was also able to reach MAE=1.74 and RMSE=13.19. In the meantime, the combined approach achieved the best result. This approach achieved RMSE=1.4 and RMSE=11.96.

The impact of different feature selection approaches is shown in Table 3. The mode without feature selection was checked in Table 1. The proposed combined ap- proach achieved MAE=3.4 and RMSE=16.92 in filter mode, respectively. In Wrap- per mode, the combined approach achieved MAE=1.32 and RMSE=10.26. Mean- while, the Autoencoder approach obtained the worst results among the feature selec- tion approaches. This approach in combination with the proposed approach, achieved MAE=2.52 and RMSE=13.16. The best results from combining the embedded and the proposed approaches were obtained. This approach was able to achieve MAE=1.22 and RMSE=9.92.

The comparison chart of different approaches is shown in Figures 3-5. As can be seen, the proposed combined approach has obtained the best results in both evaluation criteria.

Also, the error diagrams of deep learning approaches are given in Figure 6-9. According to the shape of the logarithmic graphs, overfitting and underfitting did not occur in any of the models.

Table 2: Impact of different feature selection approaches.

| Model | Feature Selection model | MAE | RMSE |
|---|---|---|---|
| LSTM | Without Feature selection | 3.30 | 12.13 |
| GRU |  | 3.34 | 12.99 |
| IndRNN |  | 1.74 | 13.19 |
| IndRNN LSTM |  | 1.4 | 11.96 |
| LSTM | Filter | 5.53 | 18.21 |
| GRU |  | 6.32 | 22.99 |
| IndRNN |  | 3.12 | 14.13 |
| IndRNN LSTM |  | 3.4 | 16.92 |
| LSTM | Wrapper | 3.56 | 13.01 |
| GRU |  | 3.57 | 13.13 |
| IndRNN |  | 1.99 | 14.29 |
| IndRNN LSTM |  | 1.32 | 10.26 |
| LSTM | Autoencoder | 6.90 | 23.21 |
| GRU |  | 7.77 | 25.09 |
| IndRNN |  | 3.09 | 16.73 |
| IndRNN LSTM |  | 2.52 | 13.16 |
| LSTM | Embedded | 3.32 | 12.01 |
| GRU |  | 3.31 | 11.65 |
| IndRNN |  | 1.72 | 12.91 |
| IndRNN LSTM |  | 1.22 | 9.92 |

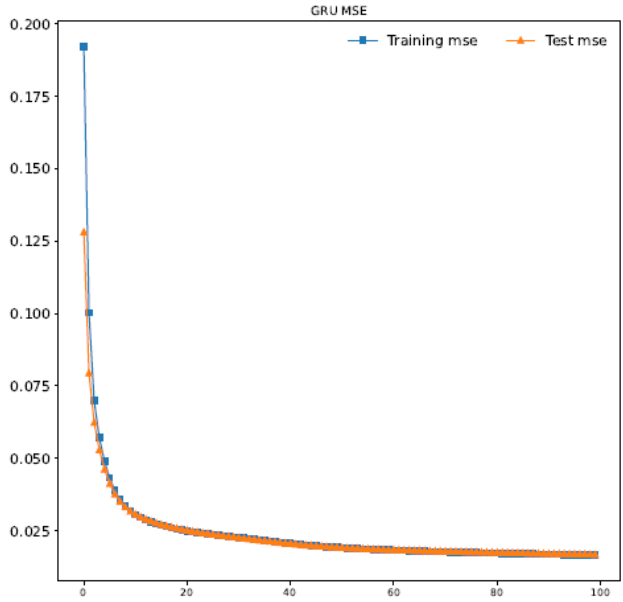

Figure 6: GRU loss

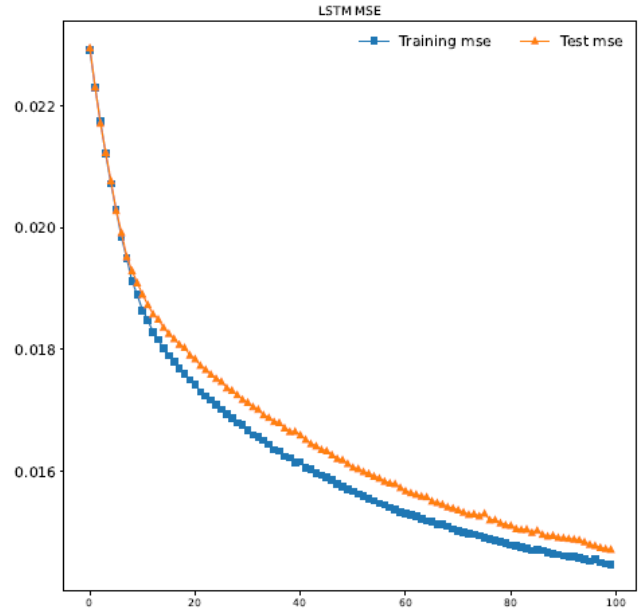

Figure 7: LSTM loss

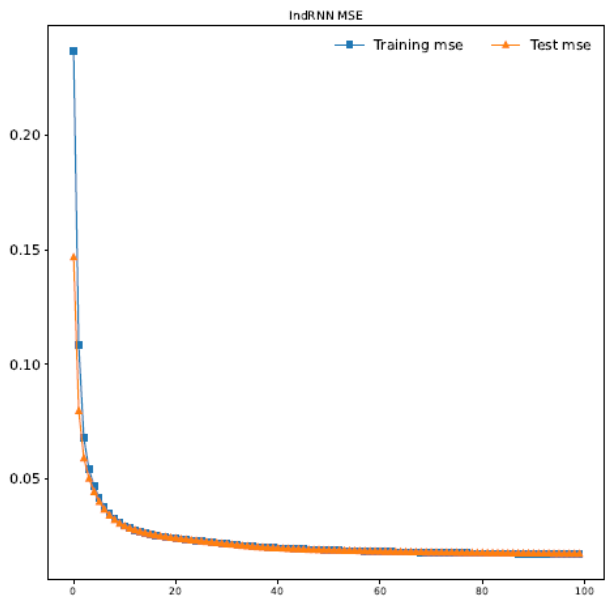

Figure 8: IndRNN loss

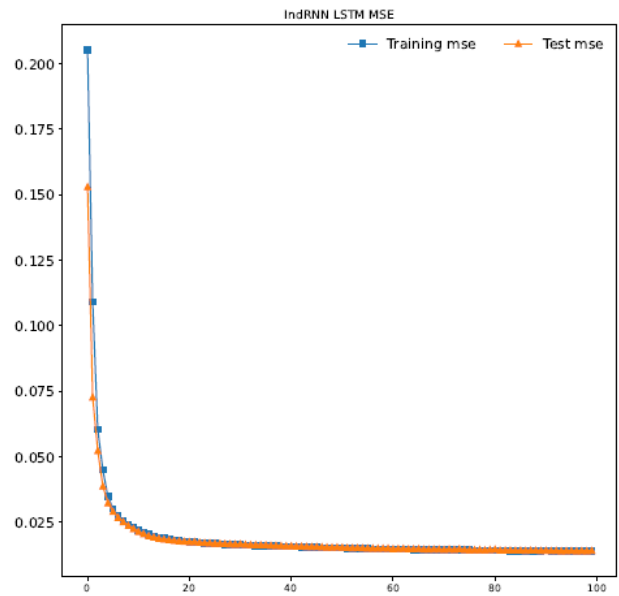

Figure 9: IndRNN + LSTM loss

## 5. Conclusion and feature works

Time series and sequence models predict future data by identifying trends in past data. The data extracted from the simulated network has the property of time series, which intensifies the trend towards the use of sequence-based models. In this article, various suggestions were presented based on feature selection approaches such as Filter, Wrapper, Embedded, and Autoencoder. Recursive deep learning approaches such as LSTM, GRU, and IndRNN were used. Also, a hybrid model that tried to learn de- pendent and non-dependent features was introduced. IndRNN LSTM hybrid approach combined with Embedded feature selection was able to achieve the lowest RMSE value on NSL-KDD data. Considering the continuous space of features of deep learning models, it is challenging to provide a single parametric space. For this purpose, the PSO algorithm can be used to find the optimal parameter space as a suggestion for future works. Also, the combination of recurrent deep learning approaches and machine learning approaches such as SVM and XGBoost can lead to improved results.